\documentclass[11pt]{article}
\usepackage{coling2018}
\usepackage{times}
\usepackage{url}
\usepackage{latexsym}
\usepackage{graphicx}
\usepackage{multicol}
\usepackage{multirow}

\title{Classifier Ensembles for Dialect and Language Variety Identification}

\author{Liviu P. Dinu\textsuperscript{1}, Alina Maria Ciobanu\textsuperscript{1}, Marcos Zampieri\textsuperscript{2}, Shervin Malmasi\textsuperscript{3} \\
  \textsuperscript{1}University of Bucharest, Romania \\
  \textsuperscript{2}University of Wolverhampton, United Kingdom \\
   \textsuperscript{3}Harvard Medical School, United States \\
  {\tt liviu.p.dinu@gmail.com} \\}

\date{}

\begin{document}
\maketitle
\begin{abstract}
In this paper we present ensemble-based systems for dialect and language variety identification using the datasets made available by the organizers of the VarDial Evaluation Campaign 2018. We present a system developed to discriminate between Flemish and Dutch in subtitles and a system trained to discriminate between four Arabic dialects: Egyptian, Levantine, Gulf, North African, and Modern Standard Arabic in speech broadcasts. Finally, we compare the performance of these two systems with the other systems submitted to the Discriminating between Dutch and Flemish in Subtitles (DFS) and the Arabic Dialect Identification (ADI) shared tasks at VarDial 2018.
\end{abstract}

\section{Introduction}
\label{intro}

Discriminating between national language varieties and dialects is an important task that is often integrated in natural language processing pipelines and applications. The problem has attracted more attention in recent years, as evidenced by a number of research papers published on languages such as English \cite{lui13}, Portuguese \cite{zampieri2016computational}, and Romanian \cite{ciobanu2016computational} and shared tasks such as tweetLID \cite{zubiaga2016tweetlid}, the PAN lab on author profiling \cite{rangel2017overview}, and the DSL shared task \cite{zampieri:2014:VarDial}.

Due to its important dialectal variation, the application of automatic dialect identification methods has been widely studied for Arabic, one of the languages we work on in this paper. A number of studies have been published on the identification of Arabic dialects and Modern Standard Arabic using user generated content (e.g. microblog and social media posts), speech transcripts, and other corpora \cite{elfardy13,zaidan14,malmasi-et-al:2015:adi,tillmann-mansour-alonaizan:2014:VarDial}. 

The other language pair we investigate in this paper is Dutch and Flemish. To the best of our knowledge, methods to discriminate between these two languages haven't been substantially investigated, a notable exception is the work by \newcite{vanderlee-vandenbosch:2017:VarDial}. Taking the scientific literature into account, these two are widely considered to be two national varieties of the same language, one variety is spoken in the Netherlands while the other is spoken in Belgium. In this paper, we refer to these two varieties as Flemish and Dutch but we acknowledge that in related work (e.g. \newcite{peirsman10}), other terms are used to define these two language varieties, such as Belgium Dutch for Flemish and Netherlandic Dutch for Dutch.\footnote{For a comprehensive survey on language and dialect identification see \newcite{jauhiainen2018automatic}.}

In this paper we present ensemble-based machine learning systems to discriminate between four Arabic dialects and Modern Standard Arabic in speech broadcasts and to discriminate between Dutch and Flemish in subtitles. We build on the experience of our previous work by improving a system that we have previously applied to similar text classification tasks such as author profiling \cite{ciobanu2017including} and native language identification \cite{zampieri-ciobanu-dinu:2017:BEA}. In our experiments, we used the datasets made available by the organizers of the Arabic, and Dutch and Flemish shared tasks of the VarDial Evaluation Campaign 2018 \cite{vardial2018report}.\footnote{\url{http://alt.qcri.org/vardial2018/index.php?id=campaign}} Finally, we compare the performance of our methods with the performance obtained by the other teams who participated in the two shared tasks.

\section{Methodology and Data}

\subsection{Data}

We used the data released by the organizers of two shared tasks of the VarDial Evaluation Campaign 2018, namely the third edition of the Arabic Dialect Identification (ADI) shared task and the first edition of the Discriminating between Dutch and Flemish in Subtitles (DFS) shared task.

The Arabic dataset made available by the organizers of the ADI shared task \cite{Ali+2016} included four Arabic dialects: Egyptian (EGY), Levantine (LEV), Gulf (GLF), North African (NOR), and Modern Standard Arabic (MSA). The data released for training and development was the same data as the data released in the 2017 edition of the VarDial evaluation campaign \cite{vardial2017report}. For testing, two new datasets were prepared: an in-domain test set and an out-of-domain dataset. The two test sets were merged and the organizers did not inform the participants that out-of-domain test data was included. In the training, development, and test sets, we were provided with acoustic features, ASR output, and phonetic features. 

The Dutch and Flemish data comes from the SUBTIEL corpus \cite{vanderlee-vandenbosch:2017:VarDial}. It consists of short excerpts of texts from subtitles of documentaries, films, and TV shows produced by a localization company that produces content for television channels in Belgium and in the Netherlands. The dataset made available by the organizers of the DFS shared task consists of 320,500 instances split into 300,000 instances for training, 20,000 for testing, and 500 for development. This amounts to a total of a little over 11 million tokens.

\subsection{Systems and Features}

We developed ensemble-based systems for dialect and language variety identification, following the methodology proposed by \newcite{malmasi-dras:2015:LT4VarDial}. The system that we propose uses multiple SVM classifiers -- each using a different type of features -- and combines their output to provide predictions. 
Such ensembles have proven to be useful for a number of related classification tasks \cite{malmasi:2016:clpsych,malmasi:2016:semltg,malmasi:2017:gdi}.

We used the Scikit-learn \cite{scikit-learn} machine learning library to implement our system. We employed the SVM implementation based on the Liblinear library \cite{liblinear}, LinearSVC\footnote{\url{http://scikit-learn.org/stable/modules/generated/sklearn.svm.LinearSVC.html}}, with a linear kernel, for the individual classifiers. We further employed the majority rule VotingClassifier\footnote{\url{http://scikit-learn.org/stable/modules/generated/sklearn.ensemble.VotingClassifier.html}} to combine the output of the SVM systems. This ensemble chooses the label that is predicted by the majority of the classifiers (which were assigned uniform weights in the ensemble). In case of ties, the ensemble chooses the label based on the ascending sort order of all labels.

We experimented with the following features, using TF-IDF weighting:

\begin{itemize}
\item Character $n$-grams, with $n$ in $\{1, ..., 8\}$;
\item Word $n$-grams, with $n$ in $\{1, 2, 3\}$;
\item Word $k$-skip bigrams, with $k$ in $\{1, 2, 3\}$.
\end{itemize}

\noindent Firstly, we trained a classifier for each type of feature. We report the individual performance of each classifier in Table~\ref{tab:results-individual}. For DFS, the best performing classifier obtained 0.701 F1 score on the development dataset, using word 3-grams as features. For ADI, the best performing classifier obtained 0.486 F1 score on the development dataset, using character 4-grams as features.

Secondly, we trained multiple ensembles, using various combinations of features, and performed a grid search to determine the optimal value for the SVM regularization parameter $C$ searching in $\{10^{-3}, ..., 10^3\}$. For DFS, the optimal $C$ value turned out to be $10^2$, and the optimal feature combination was: character n-grams with $n$ in $\{3, 4, 5, 6\}$ and word 3-grams. The best performing ensemble for DFS obtained 0.687 F1 score on the development dataset. For ADI, the optimal $C$ value turned out to be 1, and the optimal feature combination was: character n-grams with $n$ in $\{3, 4, 5\}$. The best performing ensemble for ADI obtained 0.482 F1 score on the development dataset.

\renewcommand{\tabcolsep}{1em}
\begin{table}[h]
\center
\begin{tabular}{lll}
\hline
\multirow{2}{*}{\bf Feature} & \multicolumn{2}{c}{\bf  F1 (macro)  } \\ 
 & \bf DFS & \bf ADI \\ 
\hline
Character 1-grams & 0.575 & 0.268 \\
Character 2-grams & 0.585 & 0.394 \\
Character 3-grams & 0.591 & 0.456 \\
Character 4-grams & 0.625 & \bf 0.486 \\
Character 5-grams & 0.627 & 0.466 \\
Character 6-grams & 0.653 & 0.449 \\
Character 7-grams & 0.649 & 0.433 \\
Character 8-grams & 0.645 & 0.405 \\
\hline
Word 1-grams & 0.639 & 0.451 \\
Word 2-grams & 0.663 & 0.392 \\
Word 3-grams & \bf 0.701 & 0.292 \\
\hline
Word 1-skip bigrams & 0.645 & 0.397 \\
Word 2-skip bigrams & 0.669 & 0.391 \\
Word 3-skip bigrams & 0.660 & 0.385 \\
\hline
\end{tabular}
\caption{Classification F1 score for individual classifiers on the development dataset.}
\label{tab:results-individual}
\end{table}

\section{Results}
\label{sec:results}

\subsection{Arabic Dialect Identification}

The ADI shared task 2018 is the third edition of the competition. Previous iterations were organized in 2016 \cite{dsl2016} and in 2017 \cite{vardial2017report}. Related shared tasks include last year's PAN lab on author profiling which included Arabic dialects \cite{rangel2017overview}, and the MGB-3 challenge on Arabic dialect identification \cite{ali2017speech}. 

In the 2016 edition of the ADI shared task, 18 teams submitted results in the closed submission track and four teams were ranked in the first position, taking statistical significance tests into account. The teams which were ranked first in the ADI 2016 ordered by absolute performance were: MAZA \cite{malmasi-zampieri:2016:VarDial3} which used a SVM ensemble system similar to the one we applied in this paper, UnibucKernel \cite{ionescu-popescu:2016:VarDial3} which submitted a system based on string kernels, and QCRI \cite{eldesouki-EtAl:2016:VarDial3} and ASIREM \cite{adouane-semmar-johansson:2016:VarDial3} which submitted systems based on single SVM classifiers.

In the ADI 2017, six teams submitted their system outputs and no statistical significance was calculated. The two best entries in the closed submission track of the ADI 2017 were two recurring teams, UnibucKernel \cite{ionescu-butnaru:2017:VarDial} and MAZA \cite{malmasi-zampieri:2017:VarDial2} respectively which applied adaptations of their own systems that performed well in 2016.

Next, we report the results obtained on the official test set provided by the ADI 2018 organizers. We trained our system using only the training data provided by the organizer and we compare our best system with the entries submitted to the ADI 2018. Results are presented in Table \ref{tab:adi}.

\begin{table}[!ht]
\centering
    \begin{tabular}{c l c c}
\hline
\bf Rank &    \bf Team   & \bf F1 (Macro) & System Description\\ 
\hline
1 & UnibucKernel    & 0.589 & \cite{butnaru:2018:VarDial}  \\
2 &	safina	& 0.576 & \cite{ali:arabic:2018:VarDial} \\
3 &	BZU	& 0.534 & \cite{naser:2018:VarDial} \\
3 &	SYSTRAN	& 0.529 & \cite{michon:2018:vardial} \\
3 &	T\"{u}bingen-Oslo & 0.514 & \cite{coltekin:2018:VarDial} \\
\bf 4 & \bf Best Ensemble & \bf 0.500 & \\
& Random Baseline & 0.200 & \\
\hline
\end{tabular}
\caption{ADI results. Teams were ranked taking statistical significance into account.}
\label{tab:adi}
\end{table}

The best performing system was submitted by the UnibucKernel team \cite{butnaru:2018:VarDial}, building on the experience of their previous submissions to the ADI 2016 and 2017. Our system achieved F1 score of 0.500 which significantly outperforms the baseline. It also outperforms our best individual SVM classifier (using character 4-grams as features), which achieved 0.493 F1 score on the test set. Even though the performance of our system was not much lower than the other five teams in the ADI 2018, we expected better performance from our ensemble-based system which is very similar to the entries submitted by the MAZA team which performed well in the ADI 2016 and ADI 2017. 

We did not observe the expected influence of out-of-domain data included in the test set. Including \mbox{out-of-domain} data typically makes tasks more challenging than when using only in-domain data. However, the best performance of our system in the development set was actually lower than the performance obtained on the test, 0.482 F1 score against 0.500 F1 score.

For a better understanding of our systems' performance on the test set, we present the confusion matrix in Figure \ref{fig:2}.

\begin{figure*}[!ht]
\centering
\includegraphics[width=0.65\textwidth]{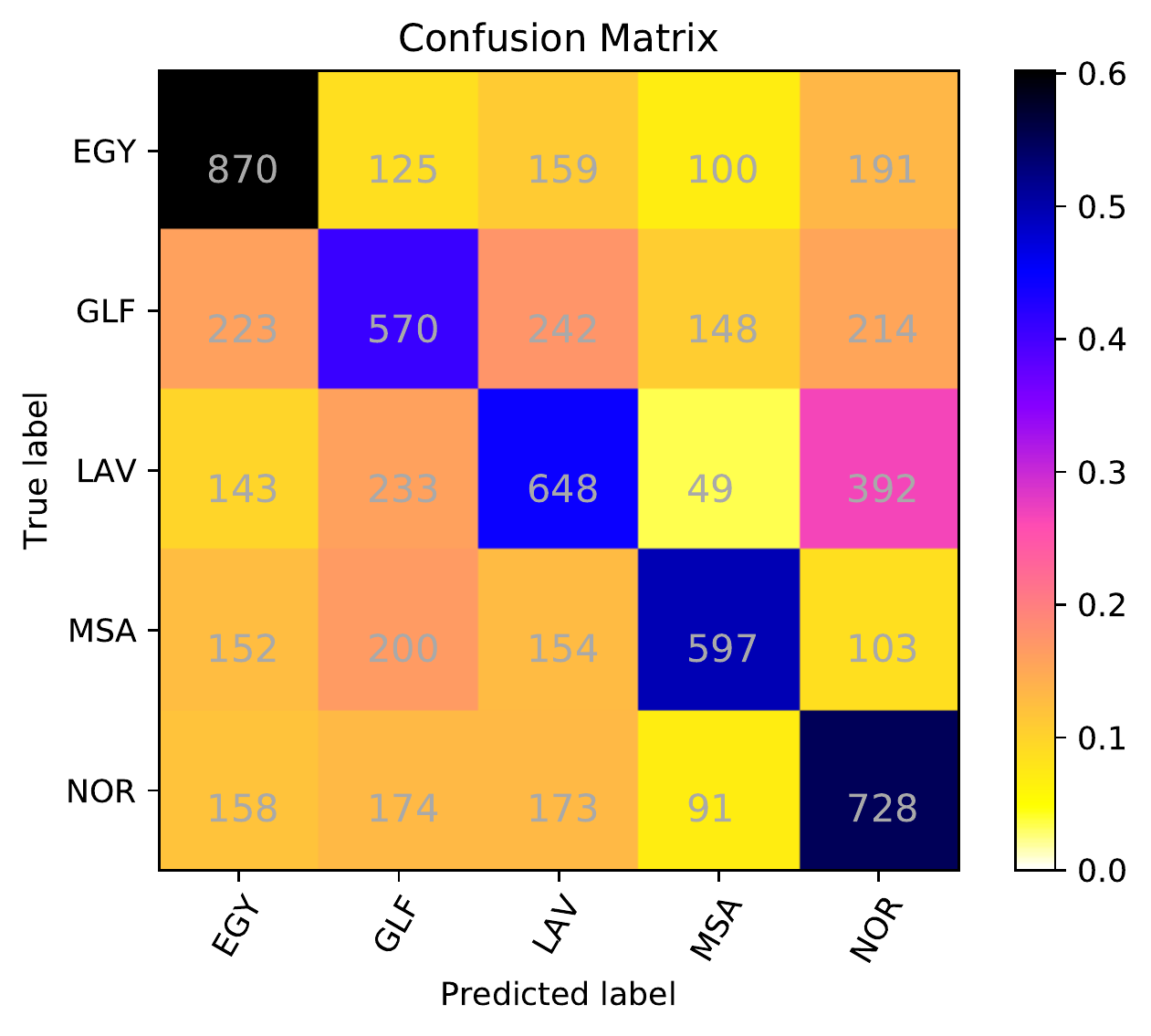}
\caption{Confusion Matrix on the ADI shared task 2018 test set.}
\label{fig:2}
\end{figure*}

\noindent We observed that our system achieved its best results identifying the Egyptian and North African dialects and the worst results identifying Gulf Arabic. The biggest confusion occurred between Levantine and North African.

\subsection{Discriminating between Dutch and Flemish}

The DFS shared task was organized for the first time in 2018. Dutch and Flemish haven't been included in the multilingual DSL shared task \cite{dsl2016} and to the best of our knowledge, no study has been published on discriminating between Dutch and Flemish with the exception of \newcite{vanderlee-vandenbosch:2017:VarDial}. This makes the results of this shared task described in \newcite{vardial2018report} very relevant for future research.

In this section, we report the results obtained on the official test set provided by the DFS shared task organizers. Our system was trained using only the training data provided by the organizer, which makes our results comparable to the results obtained by the teams who submitted their system outputs to the closed submission track. Therefore, we compare our best system with the entries submitted to the DFS shared task and we present the results of the twelve systems plus the random baseline in Table \ref{tab:adi}.

\begin{table}[!ht]
\centering
    \begin{tabular}{c l c c}
\hline
\bf Rank &    \bf Team   & \bf F1 (Macro) & System Description \\ 
\hline
1 &	T\"{u}bingen-Oslo &	0.660 & \cite{coltekin:2018:VarDial} \\
2 &	Taurus &	0.646 & \cite{vanhalteren:2018:vardial} \\
3 &	CLiPS &	0.636 & \cite{kreutz:2018:VarDial} \\
3 &	LaMa &	0.633 \\
3 &	XAC &	0.632 & \cite{barbaresi:2018:VarDial} \\
3 &	safina &	0.631 \\
4 &	STEVENDU2018 & 0.623 & \cite{du:2018:VarDial} \\
4 &	mmb\textunderscore lct & 0.620 & \cite{kroon:2018:VarDial} \\
5 &	SUKI &	0.613 & \cite{jauhianen:dfs:2018:VarDial} \\
\bf 6 & \bf	Best Ensemble & \bf 0.596 \\
7 &	dkosmajac &	0.567 \\
7 &	benf &	0.558 \\
& Random Baseline & 0.500 \\
\hline
    \end{tabular}
\caption{DFS results. Teams were ranked taking statistical significance into account.}
\label{tab:dfs}
\end{table}

\noindent Our system achieved 0.596 F1 score. It was ranked sixth in the competition, taking statistical significance into account. It outperforms the 0.500 baseline by a large margin and our best individual SVM classifier (which achieved 0.576 F1 score on the test set, using word 3-grams as features), but its performance is unfortunately not comparable to the performance obtained by the teams which were ranked in the first positions in this competition.

For a better understanding of our systems' performance in discriminating between Dutch and Flemish on the test set, we report the confusion matrix in Figure \ref{fig:2}.

\begin{figure*}[!ht]
\centering
\includegraphics[width=0.60\textwidth]{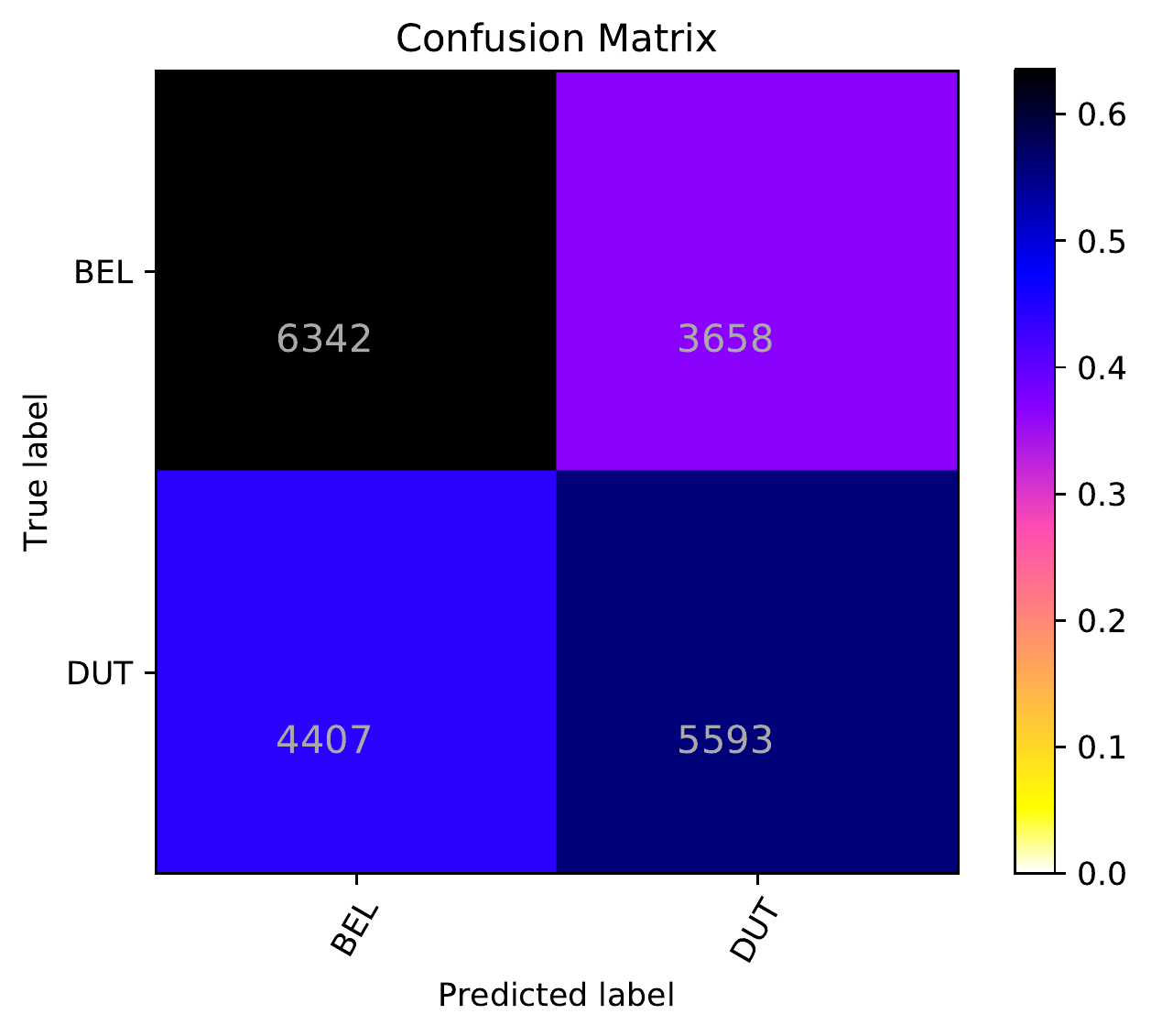}
\caption{Confusion Matrix on the DFS shared task 2018 test set.}
\label{fig:2}
\end{figure*}

\noindent We observed that our system was slightly better in identifying Flemish (BEL) than Dutch (DUT). Overall, the performance of our ensemble system is, just like for ADI, below what we would expect from a SVM ensemble-based system which, as previously stated, proved to perform well in similar shared tasks. More on that is discussed in the next section.

\section{Conclusion and Future Work}

In this paper, we present the results obtained by our ensemble-based system when discriminating between Dutch and Flemish in subtitles and when identifying dialects of Arabic using the datasets made available by the organizers of the VarDial Evaluation Campaign. We report results of 0.500 F1 score in discriminating between four Arabic dialects and MSA and 0.596 F1 score in discriminating between Dutch and Flemish.

The results obtained by our method outperform the task's baseline but we see room for improvement. For example, variations of the systems presented in this paper have been submitted to other shared tasks at VarDial 2018 achieving more competitive performance. One of the shared tasks was the Indo-Aryan Language Identification (ILI) shared task in which our system \cite{ciobanu:ili:2018:VarDial} was trained to discriminate between five closely-related languages spoken in India: Awadhi, Bhojpuri, Braj Bhasha, Hindi, and Magahi. Our system was ranked third among eight systems that competed in the task. The other shared task was the second iteration of the German Dialect Identification (GDI) shared task in which our system  \cite{ciobanu:gdi:2018:VarDial} was trained to discriminate between four (Swiss) German dialects, from Basel, Bern, Lucerne, and Zurich. In the GDI shared task our system also ranked third among eight systems.

We are currently carrying out an analysis of the most informative features learned by the classifiers and an error analysis to improve the performance of our system for future shared tasks. 

\section*{Acknowledgements}

We would like to thank the organizers of the ADI shared task and the DFS shared task for making available the datasets used in this paper.

\bibliography{vardial2018}
\bibliographystyle{acl}

\end{document}